\documentclass[12pt,a4paper,twoside]{amsart}
 \usepackage{amssymb,latexsym,amsmath,graphicx,epsf,color}
\usepackage[table]{xcolor}
\usepackage{caption}
\usepackage{subcaption}

 \numberwithin{equation}{section}

\DefineNamedColor{named}{VeryLightOrange}{rgb}{1,0.8824,0.3333}

\def \ba {\begin {eqnarray*}}
\def \ea {\end  {eqnarray*}}

\def \beq {\begin {eqnarray}}
\def \eeq {\end {eqnarray}}
\newcommand{\R}{\mathbb{R}}

\title{Photographic dataset: playing cards}

\author{Teemu Helenius, Santeri Kaupinm\"{a}ki, Samuli Siltanen and David Villacis}

\begin{document}

\maketitle
\centerline{\today}

\begin{abstract}
A photographic dataset is collected for testing image processing algorithms.
\end{abstract}

\tableofcontents

\clearpage

\section{Introduction}\label{sec:intro}

\noindent This document reports the acquisition, structure and properties of a digital photographic dataset collected at the Industrial Mathematics Laboratory of the Department of Mathematics and Statistics of University of Helsinki, Finland.\\
This dataset is designed to test and benchmark several image processing tasks, including: image denoising, image inpainting, etc., under the presence of real noise. This dataset is built with the goal of helping researchers in image processing and related areas to avoid the use of synthetic generated noise and blur to provide benchmarks and objective results.\\

\section{Materials and Methods}\label{sec:mm}

\subsection{Camera equipment} We use a PhaseOne XF medium-format camera equipped with an achromatic IQ260 digital back. The lens is Phase One Digital AF 120mm F4.
The pixel size in the resulting 16bit TIFF image file is $8964{\times}6716$.

\subsection{Lighting}

The targets were lit with five Olight X6 Marauder LED flashlights with luminous flux of 5000 lm (nominal value, 4825 lm was measured in the laboratory of the vendor www.valostore.fi). The lights were positioned at roughly equiangular arrangement. The distance of each light from the target was roughly 850 mm. The lights were heating up quickly as they were used at maximum power. Cooling was enhanced with three regular household fans.

A diffuser was placed between the lights and the target to make the lighting more uniform and to reduce sharp shadows.

\begin{figure}[ht]
\includegraphics[width=0.3\textwidth]{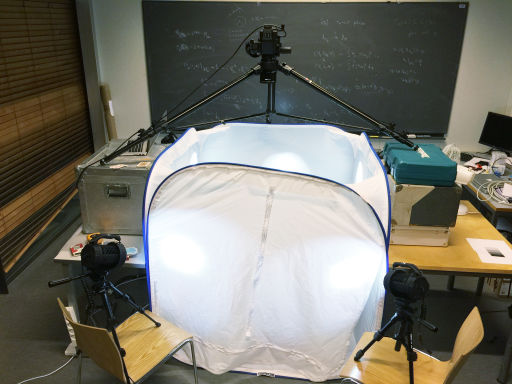}
\includegraphics[width=0.3\textwidth]{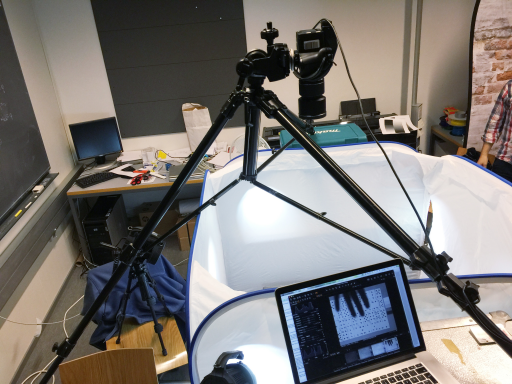}
\includegraphics[width=0.3\textwidth]{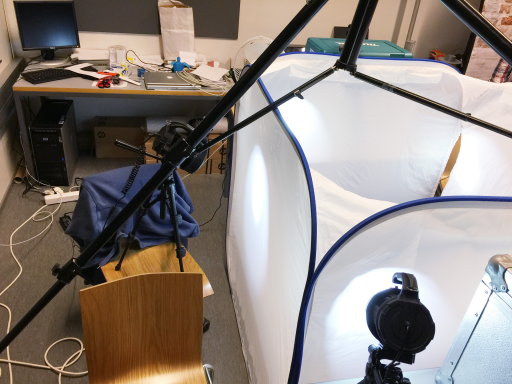}
\caption{Setup for collecting photographic data}
\label{fig:setup}
\end{figure}

See Figure~\ref{fig:setup} for the imaging setup.

\subsection{Details of the target} We placed playing cards on a horizontal surface. The camera was aimed directly down, so the optical axis was roughly vertical.

In every picture there is a five-step grayscale target for calibration.

\subsection{Varying the noise level in the data}

For each arrangement of cards, four different images were taken:
\begin{enumerate}
\item Best-quality photo (\texttt{im\_clean.tif}). Minimum ISO setting 200 and histogram approximately spanning the full dynamic range. The noise level is very small.
\item Medium-quality photo, first type (\texttt{im\_noise1.tif}). Minimum ISO setting 200 and histogram approximately spanning a quarter the full dynamic range. This will introduce round-off errors from the AD converter.
\item Medium-quality photo, second type (\texttt{im\_noise2.tif}). Maximum ISO setting 3200 and histogram approximately spanning the full dynamic range. This will introduce photon-counting noise as well as some electronic noise.
\item Low-quality photo (\texttt{im\_noise3.tif}). Maximum ISO setting 3200 and histogram approximately spanning a quarter of the full dynamic range. The noise will be increased by added round-off errors.
\end{enumerate}

\section{Calibration}\label{sec:calibration}

Due to the fact that the camera used in the acquisition process was using different parameters for different images, these images will have a different contrast and intensity values. In this section we will explain the techniques used to make the images have a common dynamic range and a similar histogram to the best-quality photo.

\subsection{Intensity Adjustment}
\begin{figure}[ht]
  \includegraphics[width=0.6\textwidth]{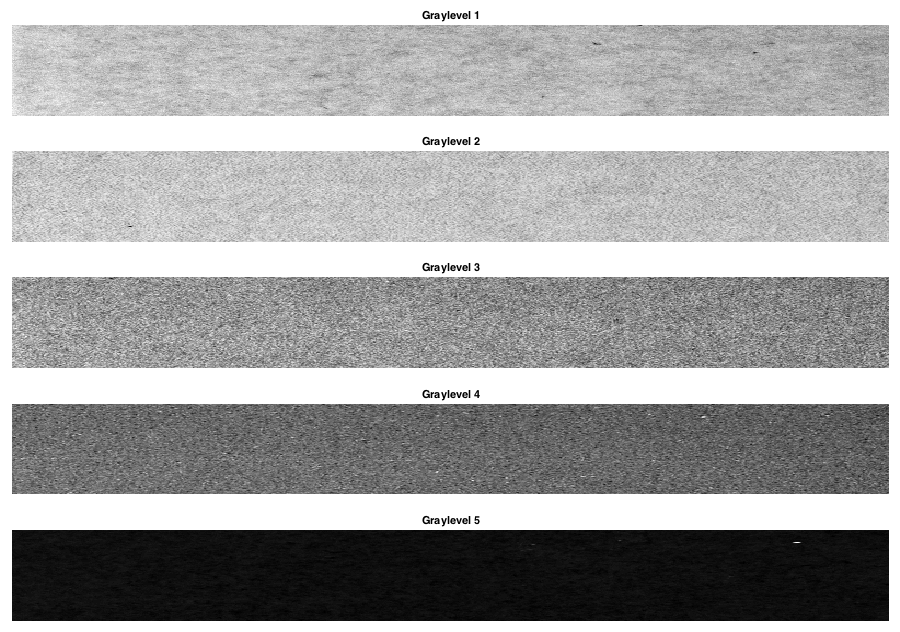}
  \caption{Calibration Grayscale Patch}
  \label{fig:Calibration Grayscale Patch}
\end{figure}

\begin{figure}[ht]
  \includegraphics[width=0.7\textwidth]{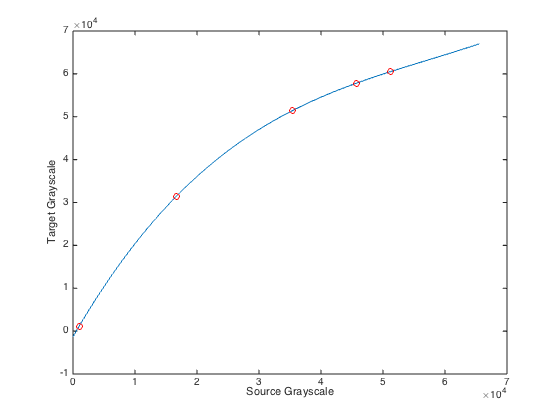}
  \caption{Intensity Adjustment Map}
  \label{fig:Intensity Adjustment Map}
\end{figure}

In this process we are interested in transforming the original image in such a way it maintains its structure while spreading the dynamic range of the image across a defined domain.\\
In order to define the grayscale levels we will make use of a grayscale calibration patch (phantom) that will be included in every image. The first step taken by the code is to extract the five graylevel areas as shown in Figure~\ref{fig:Calibration Grayscale Patch} and average the intensity values of the pixels inside it. This information is later used to build a non linear map by using a cubic spline interpolation of these graylevel calibration points, we can see an example of this mapping in Figure~\ref{fig:Intensity Adjustment Map}.

\subsection{Histogram Matching}
\begin{figure}[ht]
  \begin{subfigure}[b]{0.32\textwidth}
    \includegraphics[width=\textwidth]{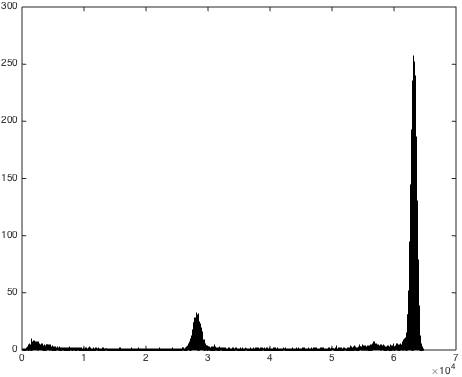}
    \caption{Clean Image}
  \end{subfigure}
  \begin{subfigure}[b]{0.32\textwidth}
    \includegraphics[width=\textwidth]{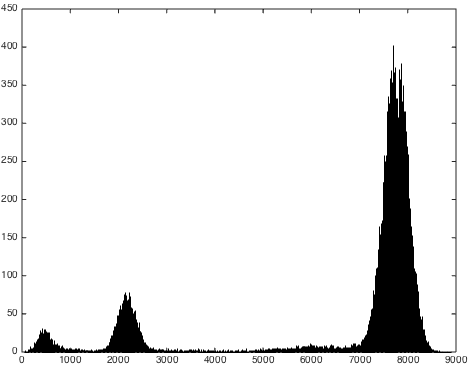}
    \caption{Original Noise 1}
  \end{subfigure}
  \begin{subfigure}[b]{0.32\textwidth}
    \includegraphics[width=\textwidth]{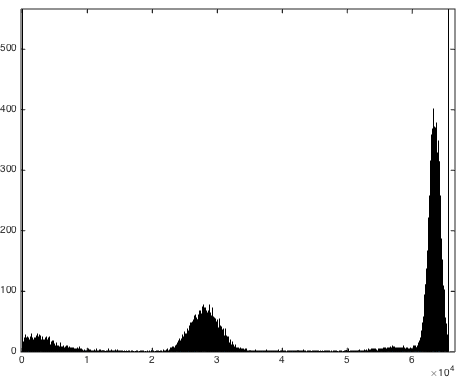}
    \caption{Corrected Noise 1}
  \end{subfigure}
  \begin{subfigure}[b]{0.32\textwidth}
    \includegraphics[width=\textwidth]{histogram_im_clean}
    \caption{Clean Image}
  \end{subfigure}
  \begin{subfigure}[b]{0.32\textwidth}
    \includegraphics[width=\textwidth]{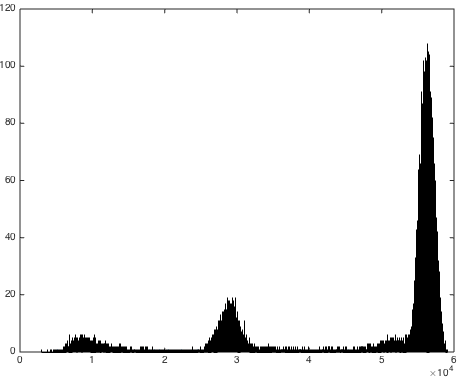}
    \caption{Original Noise 2}
  \end{subfigure}
  \begin{subfigure}[b]{0.32\textwidth}
    \includegraphics[width=\textwidth]{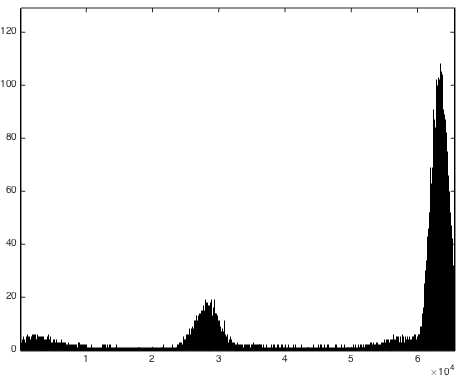}
    \caption{Corrected Noise 2}
  \end{subfigure}
  \begin{subfigure}[b]{0.32\textwidth}
    \includegraphics[width=\textwidth]{histogram_im_clean}
    \caption{Clean Image}
  \end{subfigure}
  \begin{subfigure}[b]{0.32\textwidth}
    \includegraphics[width=\textwidth]{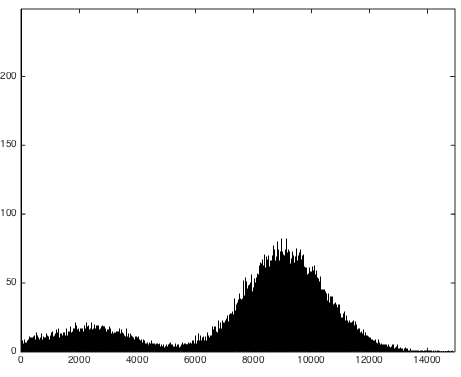}
    \caption{Original Noise 3}
  \end{subfigure}
  \begin{subfigure}[b]{0.32\textwidth}
    \includegraphics[width=\textwidth]{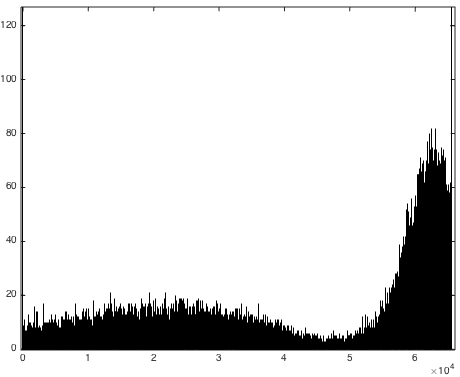}
    \caption{Corrected Noise 3}
  \end{subfigure}
  \caption{Histograms Comparison at each stage}
  \label{fig: Histograms Comparison}
\end{figure}

This procedure transforms the histogram of a givem image $A$ and a reference image $B$ and produces a modified image $\bar{A}$ such that the histograms of the images $\bar{A}$ and $B$ are similar. If we consider the graylevels in an interval $[0,65535]$, and let $r$ and $z$ denote the intensity levels for the input and output image respectively. The input levels have a probability density function $p_r(r)$ and the output levels have the \textit{specified} probability density function $p_z(z)$. A summary of the changes in the histogram of the dataset images cane be found in Figure~\ref{fig: Histograms Comparison}

\section{Results}\label{sec:results}

\begin{figure}[ht]
  \begin{subfigure}[b]{0.24\textwidth}
    \includegraphics[width=\textwidth]{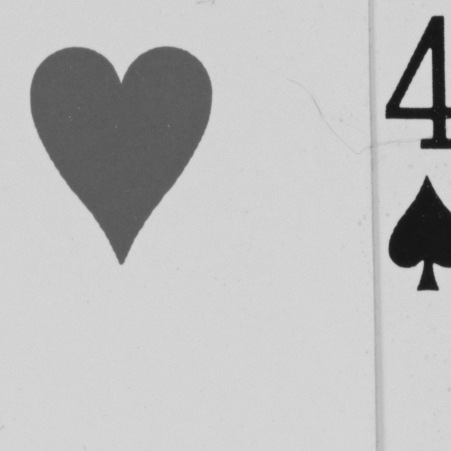}
    \caption{Clean Image}
  \end{subfigure}
  \begin{subfigure}[b]{0.24\textwidth}
    \includegraphics[width=\textwidth]{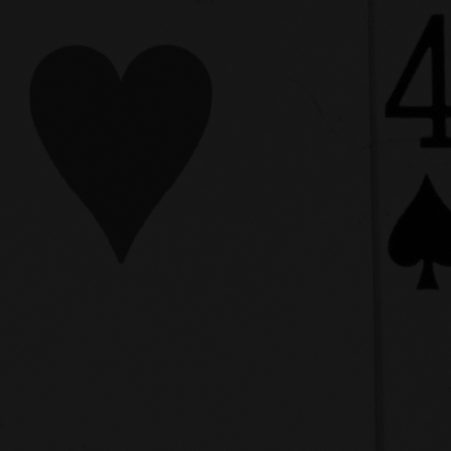}
    \caption{Noise Type 1}
  \end{subfigure}
  \begin{subfigure}[b]{0.24\textwidth}
    \includegraphics[width=\textwidth]{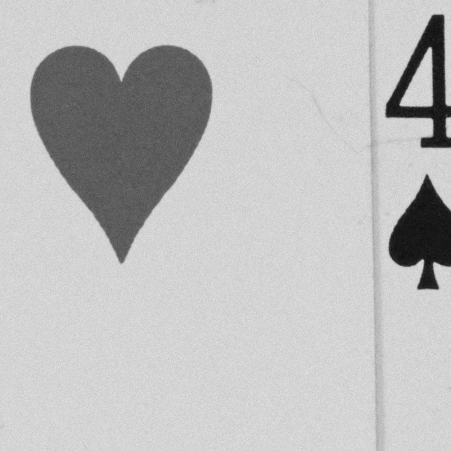}
    \caption{Noise Type 2}
  \end{subfigure}
  \begin{subfigure}[b]{0.24\textwidth}
    \includegraphics[width=\textwidth]{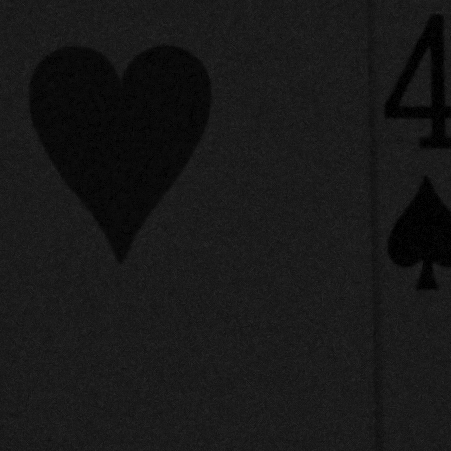}
    \caption{Noise Type 3}
  \end{subfigure}
\caption{Raw Images Comparison}
\label{fig:Raw Images Comparison}
\end{figure}

\begin{figure}[ht]
  \begin{subfigure}[b]{0.24\textwidth}
    \includegraphics[width=\textwidth]{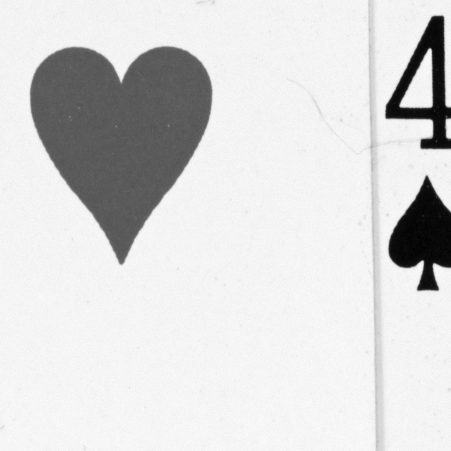}
    \caption{Clean Image}
  \end{subfigure}
  \begin{subfigure}[b]{0.24\textwidth}
    \includegraphics[width=\textwidth]{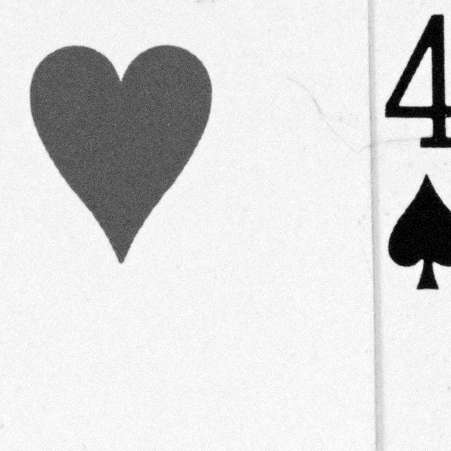}
    \caption{Noise Type 1}
  \end{subfigure}
  \begin{subfigure}[b]{0.24\textwidth}
    \includegraphics[width=\textwidth]{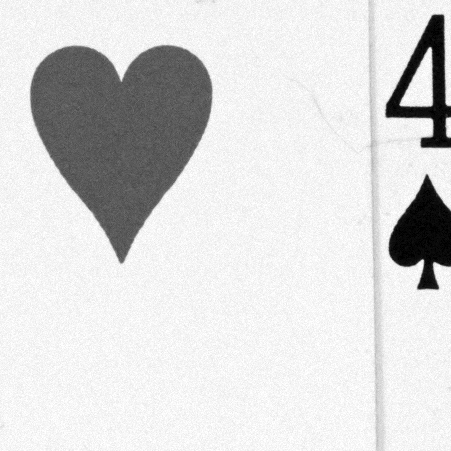}
    \caption{Noise Type 2}
  \end{subfigure}
  \begin{subfigure}[b]{0.24\textwidth}
    \includegraphics[width=\textwidth]{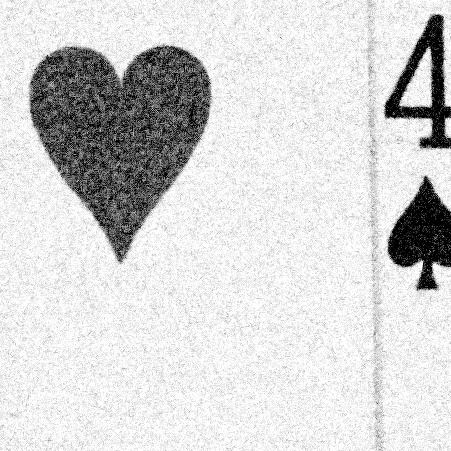}
    \caption{Noise Type 3}
  \end{subfigure}
\caption{Corrected Images Comparison}
\label{fig:Corrected Images Comparison}
\end{figure}

The lens-target distance was approximately 120 cm. The image sensor was roughly parallel with the surface of the object layer. The aperture f-stop was f/11 in all of the shots.\\
Using this non linear map described in Section~\ref{sec:calibration}, we can adjust the intensity values and histogram of the images acquired using different setup configurations, and they all will present a similar dynamic range and histogram, as an example we can see a set of raw images and it corresponding adjustment in Figure~\ref{fig:Raw Images Comparison} and a set of corrected images in Figure~\ref{fig:Corrected Images Comparison}.

\section{Discussion}\label{sec:discussion}
\begin{figure}[ht]
  \begin{subfigure}[b]{0.3\textwidth}
    \includegraphics[width=\textwidth]{corrected_im_clean}
    \caption{Clean Image}
  \end{subfigure}
  \begin{subfigure}[b]{0.3\textwidth}
    \includegraphics[width=\textwidth]{corrected_im_noise1}
    \caption{Noise Type 1}
  \end{subfigure}
  \begin{subfigure}[b]{0.3\textwidth}
    \includegraphics[width=\textwidth]{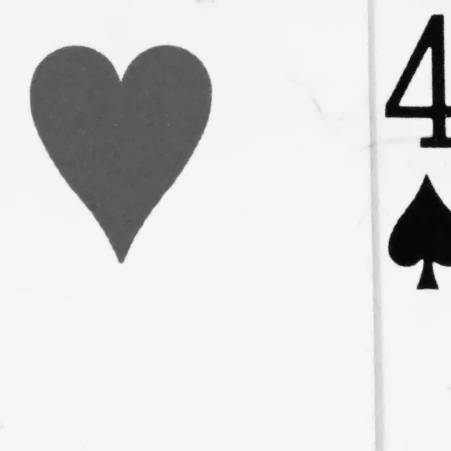}
    \caption{Denoised Image}
  \end{subfigure}
  \caption{ROF Denoising with Type 1 Noise $\lambda=1500$}
  \label{fig:ROF Denoising with Type 1 Noise}
\end{figure}

\begin{figure}[ht]
  \begin{subfigure}[b]{0.3\textwidth}
    \includegraphics[width=\textwidth]{corrected_im_clean}
    \caption{Clean Image}
  \end{subfigure}
  \begin{subfigure}[b]{0.3\textwidth}
    \includegraphics[width=\textwidth]{corrected_im_noise2}
    \caption{Noise Type 2}
  \end{subfigure}
  \begin{subfigure}[b]{0.3\textwidth}
    \includegraphics[width=\textwidth]{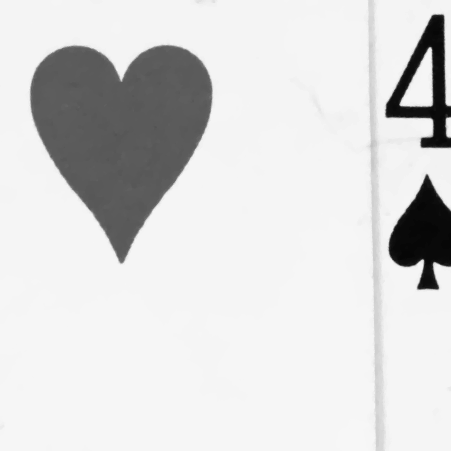}
    \caption{Denoised Image}
  \end{subfigure}
  \caption{ROF Denoising with Type 2 Noise $\lambda=1500$}
  \label{fig:ROF Denoising with Type 2 Noise}
\end{figure}

\begin{figure}[ht]
  \begin{subfigure}[b]{0.3\textwidth}
    \includegraphics[width=\textwidth]{corrected_im_clean}
    \caption{Clean Image}
  \end{subfigure}
  \begin{subfigure}[b]{0.3\textwidth}
    \includegraphics[width=\textwidth]{corrected_im_noise3}
    \caption{Noise Type 3}
  \end{subfigure}
  \begin{subfigure}[b]{0.3\textwidth}
    \includegraphics[width=\textwidth]{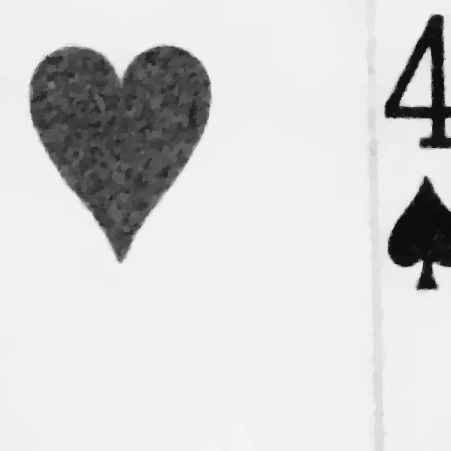}
    \caption{Denoised Image}
  \end{subfigure}
  \caption{ROF Denoising with Type 3 Noise $\lambda=10000$}
  \label{fig:ROF Denoising with Type 3 Noise}
\end{figure}

This dataset enables us to analyze the denoising task given a real world kind of noise. We can test this work with the most popular denoising models like ROF \cite{ROF} and TV-l1 \cite{TVl1}, and see its performance under real world noise conditions. Let's remember that these models make assumptions of the noise distribution, gaussian and poisson distribution for ROF and TV-l1 respectively.\\
In Figure~\ref{fig:ROF Denoising with Type 1 Noise}, Figure~\ref{fig:ROF Denoising with Type 2 Noise} and Figure~\ref{fig:ROF Denoising with Type 3 Noise} we can see the output from applying a ROF denoising model to type 1, 2 and 3 noise type respectively. In this experiment we used FISTA \cite{FISTA} algorithm to find its numerical solution. Let's recall that the ROF problem is given by:
\begin{equation}
  \min_{u\in\R^n} \| u-f \|_2^2 + \lambda \|\mathbb{K} u\|_1
\end{equation}
where $f\in\R^n$ is the given noisy image and $\mathbb{K}$ is a finite difference operator used to obtain the Total Variation norm.

\begin{figure}[ht]
  \begin{subfigure}[b]{0.3\textwidth}
    \includegraphics[width=\textwidth]{corrected_im_clean}
    \caption{Clean Image}
  \end{subfigure}
  \begin{subfigure}[b]{0.3\textwidth}
    \includegraphics[width=\textwidth]{corrected_im_noise1}
    \caption{Noise Type 1}
  \end{subfigure}
  \begin{subfigure}[b]{0.3\textwidth}
    \includegraphics[width=\textwidth]{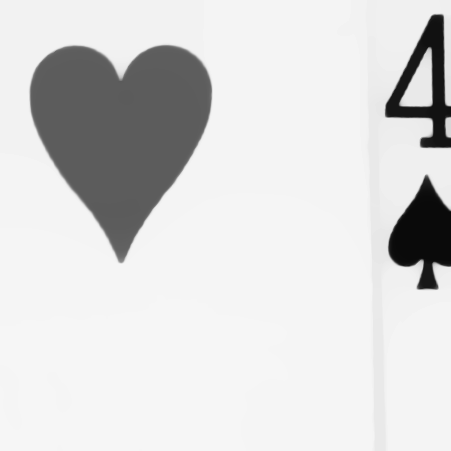}
    \caption{Denoised Image}
  \end{subfigure}
  \caption{TV-l1 Denoising with Type 1 Noise $\lambda=100$}
  \label{fig:TV-l1 Denoising with Type 1 Noise}
\end{figure}

\begin{figure}[ht]
  \begin{subfigure}[b]{0.3\textwidth}
    \includegraphics[width=\textwidth]{corrected_im_clean}
    \caption{Clean Image}
  \end{subfigure}
  \begin{subfigure}[b]{0.3\textwidth}
    \includegraphics[width=\textwidth]{corrected_im_noise2}
    \caption{Noise Type 2}
  \end{subfigure}
  \begin{subfigure}[b]{0.3\textwidth}
    \includegraphics[width=\textwidth]{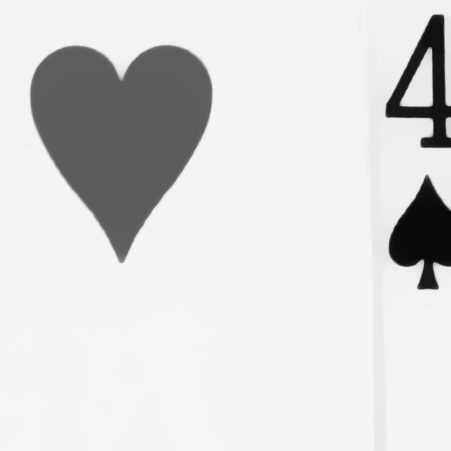}
    \caption{Denoised Image}
  \end{subfigure}
  \caption{TV-l1 Denoising with Type 2 Noise $\lambda=100$}
  \label{fig:TV-l1 Denoising with Type 2 Noise}
\end{figure}

\begin{figure}[ht]
  \begin{subfigure}[b]{0.3\textwidth}
    \includegraphics[width=\textwidth]{corrected_im_clean}
    \caption{Clean Image}
  \end{subfigure}
  \begin{subfigure}[b]{0.3\textwidth}
    \includegraphics[width=\textwidth]{corrected_im_noise3}
    \caption{Noise Type 3}
  \end{subfigure}
  \begin{subfigure}[b]{0.3\textwidth}
    \includegraphics[width=\textwidth]{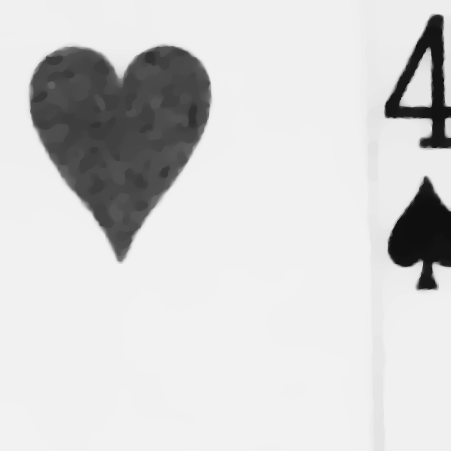}
    \caption{Denoised Image}
  \end{subfigure}
  \caption{TV-l1 Denoising with Type 3 Noise $\lambda=100$}
  \label{fig:TV-l1 Denoising with Type 3 Noise}
\end{figure}

In Figure~\ref{fig:TV-l1 Denoising with Type 1 Noise}, Figure~\ref{fig:TV-l1 Denoising with Type 2 Noise} and Figure~\ref{fig:TV-l1 Denoising with Type 3 Noise} we can see the output from applying a TV-l1 denoising model to type 1, 2 and 3 noise type respectively. In this experiment we used a Primal-Dual algorithm described in \cite{PrimalDual} algorithm to find its numerical solution. Let's recall that the TV-l1 model is given by:
\begin{equation}
  \min_{u\in\R^n} \|u-f\|_1 + \lambda \|\mathbb{K} u\|_1
\end{equation}
where $f\in\R^n$ is the given noisy image and $\mathbb{K}$ is a finite difference operator used to obtain the Total Variation norm.

\clearpage
\bibliographystyle{siam}
\bibliography{science}

\begin{thebibliography}{1}

\bibitem{FISTA}
{\sc A.~Beck and M.~Teboulle}, {\em {A Fast Iterative Shrinkage-Thresholding
  Algorithm for Linear Inverse Problems}}, SIAM Journal on Imaging Sciences, 2
  (2009), pp.~183--202.

\bibitem{PrimalDual}
{\sc A.~Chambolle, V.~Caselles, D.~Cremers, M.~Novaga, and T.~Pock}, {\em {An
  introduction to total variation for image analysis}}, Theoretical foundations
  and numerical methods for sparse recovery, 9 (2010), pp.~263--340.

\bibitem{TVl1}
{\sc M.~Nikolova}, {\em {A Variational Approach to Remove Outliers and Impulse
  Noise}}, in Journal of Mathematical Imaging and Vision, vol.~20, 2004,
  pp.~99--120.

\bibitem{ROF}
{\sc L.~I. Rudin, S.~Osher, and E.~Fatemi}, {\em Nonlinear total variation
  based noise removal algorithms}, Physica D: Nonlinear Phenomena, 60 (1992),
  pp.~259 -- 268.

\end{thebibliography}

\end{document}